%% file: main.tex
\documentclass{article}

\usepackage{PRIMEarxiv}

\usepackage[utf8]{inputenc}    % For UTF-8 encoding
\usepackage[T1]{fontenc}       % For font encoding
\usepackage{hyperref}          % For hyperlinks
\usepackage{url}               % For URL formatting
\usepackage{booktabs}          % For professional-quality tables
\usepackage{amsfonts}          % For additional math fonts (if needed)
\usepackage{nicefrac}          % For compact fractions (optional)
\usepackage{microtype}         % For microtypography improvements
\usepackage{graphicx}          % For including images
\usepackage{tabularx}          % For flexible width tables
\usepackage{fancyhdr}          % For custom headers
\usepackage{lipsum}            % For placeholder text (optional, remove if unnecessary)
\usepackage{amsmath}
\usepackage{multirow}
\usepackage{multicol}

\usepackage{tikz}
\usepackage{pifont}
\usepackage[table]{xcolor} % for table colors
\usepackage{pifont}        % for \ding symbols
\usepackage{array}

\graphicspath{{media/}}     % organize your images and other figures under media/ folder

\title{Stresnet \& Styolo : A New Family of Compact Classification and Object Detection Models for MCUs} 

\author{
  {\bf Sudhakar Sah, Ravish Kumar} \\ \\
  STMicroelectronics\\
  Toronto, Canada \\
  \texttt{sudhakar.sah@st.com}
}
\begin{document}

\maketitle

\begin{abstract}
Recent advancements in lightweight neural networks have significantly improved the efficiency of deploying deep learning models on edge hardware. However, most existing architectures still compromise accuracy for latency, which limits their applicability on MCU/NPU-based devices. In this work, we introduce two new model families — STResNet for image classification and STYOLO for object detection — jointly optimized for accuracy, efficiency, and memory footprint on resource-constrained platforms. The proposed STResNet series (ranging from Nano to Tiny variants) achieves competitive ImageNet-1K accuracy within a 4M parameter budget. Specifically, STResNetMilli attains 70.0\% Top-1 accuracy with only 3.0M parameters, outperforming MobileNetV1 and ShuffleNetV2 at comparable computational complexity. For object detection, STYOLOMicro and STYOLOMilli achieve 30.5\% and 33.6\% mAP, respectively, on the MS-COCO dataset, surpassing YOLOv5n and YOLOX-Nano in both accuracy and efficiency. Furthermore, when STResNetMilli is used as a backbone with the Ultralytics detection head, it approaches the performance of the YOLOv11n model under the latest Ultralytics training environment.

\end{abstract}

\section{Introduction}
The growing adoption of edge intelligence has intensified the demand for compact and efficient neural networks capable of operating within the stringent memory and compute limits of resource-constrained hardware such as Microcontroller Units (MCUs) and Neural Processing Units (NPUs). Conventional Convolutional Neural Networks (CNNs), such as ResNet~\cite{he2016deep}, while highly accurate, are often impractical for deployment on such platforms due to their substantial computational and memory requirements.

Lightweight architectures such as MobileNet~\cite{howard2017mobilenets}, ShuffleNet~\cite{zhang2018shufflenet}, EfficientNet~\cite{tan2019efficientnet}, SqueezeNet~\cite{iandola2016squeezenet}, and NASNet~\cite{zoph2018nasnet} have been proposed to address these challenges. However, they frequently rely on specialized operations—such as depthwise separable convolutions, fire modules, channel shuffle, and squeeze–excitation blocks—that are often unsupported or inefficient on MCU/NPU hardware. Moreover, these operations can be less amenable to quantization~\cite{jacob2018quantization}, further complicating hardware deployment.

To overcome these limitations, extensive research has focused on model compression and architectural optimization techniques, including pruning~\cite{han2015deep}, quantization~\cite{jacob2018quantization}, low-rank decomposition~\cite{denton2014exploiting}, and neural architecture search (NAS)~\cite{elsken2019neural, wu2019fbnet}, aiming to create models that are both memory- and compute-efficient while maintaining competitive accuracy on embedded platforms.

In this work, we introduce a new family of ultra-compact classification and detection models—specifically designed for MCU and NPU deployment—by combining layer decomposition~\cite{kim2015compression, lebedev2014speeding} with Neural Architecture Search (NAS) in a unified framework termed \emph{CompressNAS}. The proposed classification backbone, \emph{STResNet}, is a decomposed variant of ResNet~\cite{he2016deep} that achieves between \textbf{3$\times$ and 12$\times$ compression} while preserving competitive accuracy.

The \emph{STResNet} family adopts a clean and hardware-efficient design that facilitates seamless deployment on low-power devices. It retains the fundamental residual block structure of ResNet~\cite{he2016deep} but applies \textbf{layer decomposition} and \textbf{channel compression} to substantially reduce memory and compute requirements. Unlike complex NAS-generated or heavily engineered tiny architectures, STResNet relies solely on standard convolutional operations, resulting in improved numerical stability, quantization compatibility, and predictable behavior across embedded platforms.

Despite its structural simplicity, \emph{STResNet} achieves performance competitive with state-of-the-art handcrafted lightweight models, demonstrating that a carefully decomposed ResNet backbone can effectively match or surpass architectures such as MobileNet~\cite{howard2017mobilenets, sandler2018mobilenetv2, howard2019mobilenetv3} and EfficientNet~\cite{tan2019efficientnet}—all while maintaining a more deployment-friendly design. 

Building upon the lightweight \emph{STResNet} classification backbone, we extend our design to object detection by integrating the decomposed ResNet architecture into a YOLOX-style framework~\cite{ge2021yolox}, resulting in a new family of efficient detectors termed \emph{STYOLO}. Modern one-stage detectors such as YOLOv5~\cite{jocher2020yolov5}, YOLOv8~\cite{ultralytics2023yolov8}, YOLOX~\cite{ge2021yolox}, and YOLOv11~\cite{yolo11_ultralytics} are typically trained end-to-end on large-scale datasets like MS-COCO~\cite{lin2014microsoft}. However, such training strategies often overlook the benefits of employing specialized, pretrained backbones that are optimized for low-resource hardware.

In our framework, the \emph{STYOLO} detector incorporates the ImageNet-pretrained, compressed \emph{STResNet} backbone into the YOLOX detection head and neck. Unlike conventional pipelines that initialize backbones with random weights, this strategy leverages a pretrained, hardware-optimized backbone to achieve faster convergence, better feature reuse, and higher mAP scores under stringent model size constraints.

Empirical evaluations on the STMicroelectronics \emph{STM32N6 Neural Art NPU} demonstrate that \emph{STYOLO} achieves higher mAP on the MS-COCO dataset than YOLOv5n and approaches the accuracy of YOLOv8n, while maintaining a comparable model footprint. These results validate that the proposed modular design and pretrained backbone strategy offer a compelling trade-off between accuracy and efficiency, enabling superior embedded object detection performance compared to conventional end-to-end training approaches.

The key contributions of this paper are summarized as follows:
\begin{itemize}
    \item We propose a novel \textbf{STResNet} family—an extremely compact classification model that combines layer decomposition and NAS-based channel optimization, achieving 3–12$\times$ compression with minimal accuracy degradation.
    \item We introduce \textbf{STYOLO}, a detection framework that integrates the decomposed ResNet backbone into YOLOX, achieving competitive accuracy and efficiency on MCU- and NPU-class hardware.
    \item We develop a \textbf{training strategy} that leverages a pre-trained and compressed STResNet backbone for initializing the STYOLO detector, enabling faster convergence and improved performance compared to end-to-end training from scratch.
    \item We perform \textbf{extensive experiments and real-hardware benchmarks} on the latest STMicroelectronics STM32N6, demonstrating that STYOLO outperforms YOLOv5n and approaches YOLOv8n performance for similar model sizes.
\end{itemize}

\section{Related Work}

\subsection{Lightweight and Tiny Deep Learning Models}

The increasing demand for on-device intelligence has led to substantial research into lightweight and compact neural network architectures designed for edge devices. Early works such as SqueezeNet~\cite{iandola2016squeezenet} and MobileNet~\cite{howard2017mobilenets, sandler2018mobilenetv2} introduced efficient convolutional designs that significantly reduced parameter counts and floating-point operations (FLOPs) without major accuracy losses. These models popularized techniques such as \textbf{depthwise separable convolutions}, pointwise projections, and bottleneck expansions, which became standard components in modern efficient architectures.  
   
ShuffleNet~\cite{zhang2018shufflenet} further improved efficiency through channel shuffling, while EfficientNet~\cite{tan2019efficientnet} introduced a compound scaling method that uniformly balances depth, width, and resolution. Despite their efficiency, such handcrafted models often rely on specialized layers types that are not always hardware-friendly for quantization or low-level deployment on MCUs and NPUs. In particular, depthwise convolutions, while computationally efficient on GPUs, can lead to degraded throughput or instability when quantized to low-bit formats~\cite{jacob2018quantization}.

\subsection{Neural Architecture Search (NAS)}

The advent of Neural Architecture Search (NAS) provided an automated approach to design efficient models optimized for specific hardware. NASNet~\cite{zoph2018learning} pioneered reinforcement learning-based search strategies, while MnasNet~\cite{tan2019mnasnet} introduced multi-objective optimization to jointly consider latency and accuracy. FBNet~\cite{wu2019fbnet} and ProxylessNAS~\cite{cai2019proxylessnas} further advanced this direction by introducing differentiable NAS frameworks that incorporate hardware constraints directly into the search process.  

While these NAS-based architectures achieve state-of-the-art performance on mobile platforms, they often result in highly fragmented or irregular layer topologies that complicate deployment on resource-limited MCUs and NPUs. Our approach, in contrast, focuses on structural simplicity by retaining a ResNet-style topology while using NAS only for channel and decomposition-level optimization, thereby preserving deployment consistency across hardware backends.

\subsection{Model Compression  Techniques}

Model compression through pruning, quantization, and low-rank factorization has been a parallel strategy to reduce inference complexity. Han \textit{et al.}~\cite{han2015deep} introduced deep compression through iterative pruning and quantization, while Denton \textit{et al.}~\cite{denton2014exploiting} and Lebedev \textit{et al.}~\cite{lebedev2014speeding} demonstrated the use of tensor decomposition (CP and Tucker) for accelerating convolutional layers. Subsequent work by Kim \textit{et al.}~\cite{kim2015compression} optimized low-rank decomposition for mobile hardware, achieving substantial performance gains with minimal accuracy loss.

Our proposed method draws inspiration from these decomposition-based approaches but integrates them within a NAS-guided pipeline, allowing both the decomposition rank and layer configuration to be optimized jointly. This hybrid design leads to compact models that maintain representational power while remaining efficient and hardware-friendly.

In summary, prior works have shown that lightweight model design, NAS, and decomposition can independently yield compact architectures. However, few approaches have combined these paradigms in a unified framework specifically tailored for MCU/NPU deployment. Our proposed \textbf{CompressNAS-ResNet} and \textbf{STYOLO} frameworks bridge this gap by combining decomposition-driven compression with NAS-based channel optimization, all within a simple ResNet-style topology that ensures compatibility, stability, and high efficiency on embedded inference hardware.

\section{CompressNAS}
\label{sec:CompressNAS}
CompressNAS is an architectural optimization framework that integrates layer decomposition with a NAS-guided channel optimization strategy. Specifically, Tucker decomposition is applied to each layer in the network, where the optimal rank for every layer is determined. The selection of these ranks is formulated as a global optimization problem, capturing the interdependence of decomposition choices across layers.
The impact of decomposing each layer is independently assessed in terms of accuracy degradation ($\Delta$acc) and memory footprint reduction (or flash size, $\Delta$flash). Subsequently, an Integer Linear Programming (ILP)-based search algorithm is employed to identify the optimal rank configuration for all layers, subject to predefined hardware constraints.

\begin{figure}
    \centering
    \includegraphics[width=\textwidth]{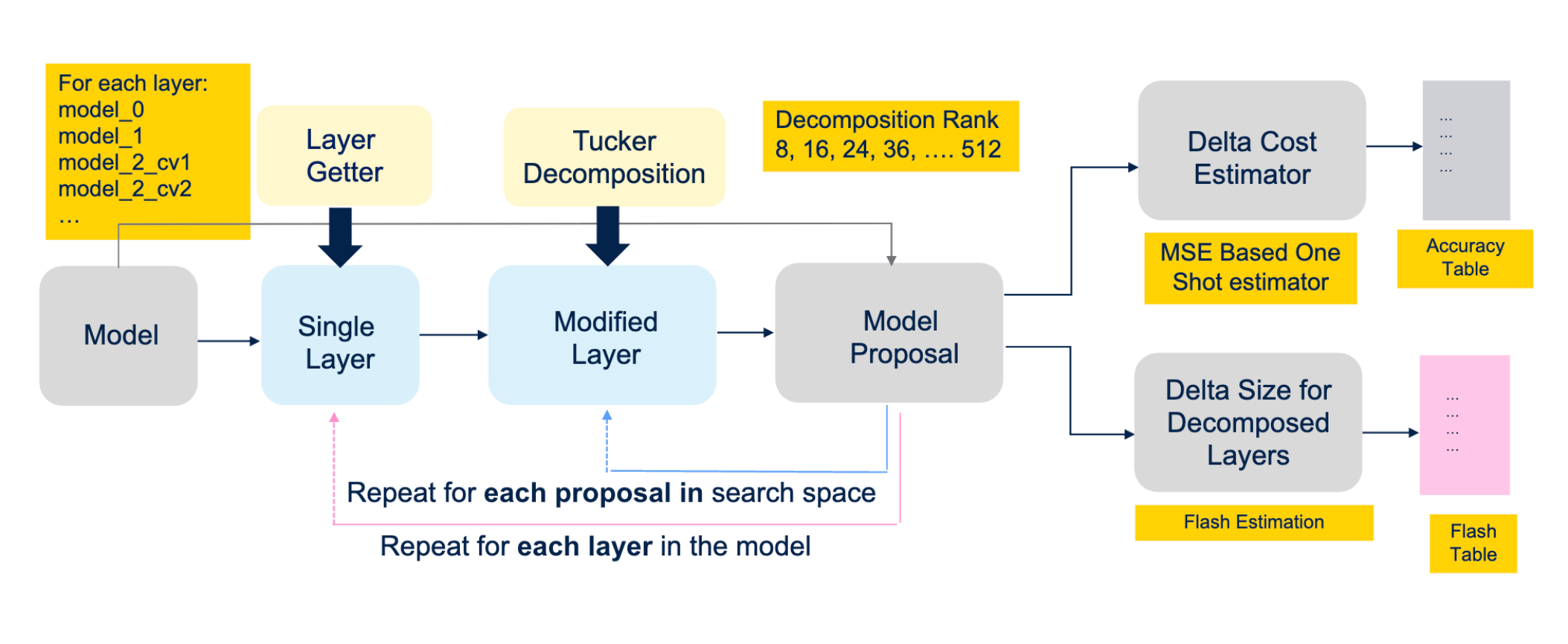}
    \caption{\emph{CompressNAS}: Model Proposal Generation and Profiling}
    \label{fig:quark_proposal_generation}
\end{figure}

\subsection{Network Proposals}
Figure \ref{fig:quark_proposal_generation} illustrates the complete CompressNAS methodology. For convolutional layers, the number of decomposition proposals is determined by the number of output channels. An exhaustive search strategy is used to identify the optimal rank by generating multiple rank proposals, starting from 8 channels and incrementing in configurable steps of 8 (or 4), depending on the desired search granularity.
For each rank proposal, Tucker decomposition is applied to the corresponding layer, and the decomposed layer replaces the original in the main architecture. The modified model is then evaluated to measure the resulting changes in accuracy and flash memory usage, with these results recorded in the corresponding lookup tables for subsequent ILP-based optimization.
\subsection{Accuracy Estimator}
After layer replacement, the model can be retrained or fine-tuned on the target dataset to estimate its final accuracy. However, given that the decomposition process can generate hundreds or even thousands of model proposals, retraining each configuration is computationally infeasible. To address this, Zero-Cost (ZC) proxies are employed, which estimate the impact of each layer modification using only a single forward pass through the altered architecture.

Although several ZC proxies were evaluated, their predictions did not align with the expected theoretical trend — i.e., higher ranks yielding lower reconstruction errors. Consequently, a Mean Squared Error (MSE)-based proxy was adopted, which computes the error between the output tensors of the modified and reference layers. Unlike traditional ZC proxies, the MSE-based approach consistently demonstrates the expected correlation between rank and error, providing a more reliable performance estimate.
\subsection{Flash Estimator}
Each modified model is exported to the ONNX format for consistent evaluation and hardware-level analysis. The difference in model size between the modified and original architectures is computed using Equation \ref{eq:delta_flash}, where $M$ denotes the number of output channels, $N$ represents the input channels, $k$ corresponds to the kernel size, and $R$ indicates the decomposed rank. This formulation quantifies the change in memory footprint ($\Delta$flash) resulting from layer decomposition.
\begin{equation}
\label{eq:delta_flash}
\Delta flash = N M k^2 - \left(NR\cdot 1\!\times\!1 + R^2 \cdot 3\!\times\!3 + RM\cdot 1\!\times\!1 \right)
\end{equation}
\subsection{Neural Architecture Search}
After constructing the lookup tables, an Integer Linear Programming (ILP)-based search algorithm is employed to determine the optimal architecture configuration that satisfies the predefined hardware constraints, as formulated in Equation \ref{eq:ilp_formulation}. This ILP optimization ensures an efficient trade-off between accuracy and memory footprint, yielding a decomposed architecture that best fits the target device specifications.
\begin{equation}
\label{eq:ilp_formulation}
\begin{aligned}
Accuracy = \max \quad & \sum_{(i,j) \in E} \Delta \text{accuracy}_{ij} \\
\text{s.t.} \quad & \sum_{(i,j) \in E} \Delta \text{flash}_{ij} \leq flash_{\max}.
\end{aligned}
\end{equation}

After varying the constraints on accuracy and flash, two optimized model families are formed: \emph{STResNet} and \emph{STYOLO}.

\section{Architecture: STResNet}
\label{sec:stresnet}

The \emph{STResNet} family comprises a series of decomposed variants of the ResNet architecture, designed to achieve different trade-offs between model complexity and accuracy. Five configurations are introduced—\emph{STResNet-Tiny}, \emph{STResNet-Milli}, \emph{STResNet-Micro}, \emph{STResNet-Nano}, and \emph{STResNet-Pico}—each corresponding to a different level of decomposition and resulting parameter count. All variants contain fewer than 4 million parameters, with \emph{STResNet-Tiny} being the largest and most accurate model, and \emph{STResNet-Pico} being the most compact variant optimized for extreme resource constraints.

\subsection{Simplified Architecture}

The \emph{STResNet} family consists of a series of decomposed variants of the ResNet architecture, each designed to balance model complexity, memory footprint, and accuracy. Five configurations are introduced—\emph{STResNet-Tiny}, \emph{STResNet-Milli}, \emph{STResNet-Micro}, \emph{STResNet-Nano}, and \emph{STResNet-Pico}—corresponding to progressively higher levels of decomposition and reduced parameter counts. All variants contain fewer than 4 million parameters, with \emph{STResNet-Tiny} serving as the largest and most accurate configuration, and \emph{STResNet-Pico} representing the most compact model optimized for deployment under extreme resource constraints. This hierarchical scaling enables flexible deployment across a wide range of MCU and NPU hardware, depending on available compute and memory budgets. 

\subsection{Flash Consideration}

Flash memory constraints are a critical factor when deploying CNN models on MCUs and edge NPUs. The decomposed layers in STResNet significantly reduce the number of parameters that must be stored in non-volatile memory. Each convolutional layer is represented using three smaller factorized matrices, which collectively maintain representational capacity while lowering storage requirements. In addition, the absence of specialized or irregular layers removes the need for custom kernels, enabling shared, uniform convolution implementations that are both memory-efficient and cache-friendly.

\subsection{RAM Consideration}

Runtime memory usage, particularly for feature map storage, poses a major bottleneck in embedded inference. While the decomposed layers in STResNet do not inherently reduce RAM consumption, analysis revealed that a sub-layer within the stem block accounted for the highest memory utilization. To address this, a projection layer was introduced, as described in Section~\ref{subsec:projection_layer}. This modification achieved approximately a $2\times$ reduction in RAM usage with a negligible accuracy drop of less than $0.5\%$.

\subsection{Latency Consideration}

Inference latency on NPUs and MCUs is influenced not only by computational complexity (FLOPs) but also by factors such as operator fusion, kernel regularity, and memory bandwidth. The STResNet architecture preserves high operator regularity by exclusively employing standard $3\times3$ and $1\times1$ convolutions across all layers, thereby facilitating efficient hardware acceleration without incurring the overhead associated with diverse or irregular convolution types.

In summary, STResNet demonstrates that a decomposition-based design, when carefully optimized, can yield MCU- and NPU-efficient models exhibiting low latency, high flash efficiency, and optimized RAM utilization—all achieved without relying on depthwise separable convolutions or manually crafted modules.

% \subsection{Design Philosophy} STResNet is designed around three core ideas: \begin{itemize} \item \textbf{Multi-branch Training, Single-branch Inference:} During training, each block contains multiple branches (identity, $1\times1$ conv, $3\times3$ conv) that are merged into a single $3\times3$ equivalent kernel for inference. \item \textbf{Residual Efficiency:} Residual shortcuts enable stable training even in tiny models while minimizing parameter overhead. \item \textbf{Scalable Width and Depth:} Each variant adjusts channels and layers, enabling deployment from smartphones to microcontrollers. \end{itemize}

\begin{table*}[h!]
\centering
\small
\caption{ResNet-18 Architecture }
\label{tab:resnet18_ref_arch}
\begin{tabular}{l c c}
\hline
\textbf{Block / Layer} & \textbf{In $\to$ Out} & \textbf{Conv Layers (Inner Channels)} \\
\hline
\textbf{Stem} & 3 $\to$ 64 & 7x7 /3→64 /2, MaxPool 3x3 /2 \\
\hline
\textbf{Layer1} & 64 $\to$ 64 & Block1: 3x3/64, 3x3/64; Block2: 3x3/64, 3x3/64 \\
\hline
\textbf{Layer2} & 64 $\to$ 128 & Block1: 3x3/64 s2, 3x3/128; Block2: 3x3/128, 3x3/128 \\
\hline
\textbf{Layer3} & 128 $\to$ 256 & Block1: 3x3/128 s2, 3x3/256; Block2: 3x3/256, 3x3/256 \\
\hline
\textbf{Layer4} & 256 $\to$ 512 & Block1: 3x3/256 s2, 3x3/512; Block2: 3x3/512, 3x3/512 \\
\hline
\textbf{FC} & 512 $\to$ 1000 & Linear Layer \\
\hline
\end{tabular}
\end{table*}

\begin{table*}[h!]
\centering
\small
\caption{STResNet-Nano Architecture }
\label{tab:stresnet_nano_arch}
\begin{tabular}{l c c}
\hline
\textbf{Block / Layer} & \textbf{In $\to$ Out} & \textbf{Conv Layers (Inner Channels)} \\
\hline
\textbf{Stem} & 3 $\to$ 64 & 1x1/3→3, 7x7/3→8, 1x1/8→32 \\ & & \textbf{proj.1x1/32→64} \\
\hline
\textbf{Layer1} & 64 $\to$ 64 & Block1: 1x1/64→32, 3x3/32, 1x1/32→64; 1x1/64→16, 3x3/16, 1x1/16→64 \\
& & Block2: 1x1/64→40, 3x3/40, 1x1/40→64; 1x1/64→8, 3x3/8, 1x1/8→64 \\
\hline
\textbf{Layer2} & 64 $\to$ 128 & Block1: 1x1/64→32, 3x3/32 s2, 1x1/32→128; 1x1/128→8, 3x3/8, 1x1/8→128 \\
& & Block2: 1x1/128→64, 3x3/64, 1x1/64→128; 1x1/128→16, 3x3/16, 1x1/16→128 \\
\hline
\textbf{Layer3} & 128 $\to$ 256 & Block1: 1x1/128→48, 3x3/48 s2, 1x1/48→256; 1x1/256→16, 3x3/16, 1x1/16→256 \\
& & Block2: 1x1/256→48, 3x3/48, 1x1/48→256; 1x1/256→8, 3x3/8, 1x1/8→256 \\
\hline
\textbf{Layer4} & 256 $\to$ 512 & Block1: 1x1/256→32, 3x3/32 s2, 1x1/32→512; 1x1/512→8, 3x3/8, 1x1/8→512 \\
& & Block2: 1x1/512→48, 3x3/48, 1x1/48→512; 1x1/512→8, 3x3/8, 1x1/8→512 \\
\hline
\end{tabular}
\end{table*}

% \begin{figure}[h] \centering \includegraphics[width=0.8\linewidth]{stresnet_block.png} \caption{STResNet residual block with structural re-parameterization (training vs. inference).} \label{fig:stresnet_block} \end{figure}

\section{Architecture : STYOLO} 
\label{sec:styolo}
The STYOLO family of object detectors is built upon the STResNet backbone, incorporating a modified neck and detection head derived from the YOLOX architecture. The overall training pipeline follows the YOLOX framework, with several targeted modifications aimed at improving efficiency and accuracy. In alignment with the size-based hierarchy of STResNet, five variants of STYOLO are introduced—STYOLO-Tiny, STYOLO-Milli, STYOLO-Micro, STYOLO-Nano, and STYOLO-Pico—each corresponding to a specific backbone configuration. This section details the architectural adjustments and training strategies employed to enhance the performance of the STYOLO model family across diverse resource constraints.

\subsection{Neck Adjustment}
In the STYOLO architecture, the backbone generates raw multi-scale feature maps—\texttt{dark3}, \texttt{dark4}, and \texttt{dark5}—that possess relatively high channel dimensions. These feature maps are not directly compatible with the YOLOX-style neck, which expects reduced channel sizes for efficient multi-scale feature aggregation. To ensure proper alignment, a channel projection is applied using $1 \times 1$ convolutions that compress the feature dimensions before they are fed into the neck (Table~\ref{tab:neck_alignment}). Specifically, backbone outputs of 128, 256, and 512 channels are projected to 64, 128, and 256 channels, respectively. This projection preserves spatial resolution while significantly reducing computational overhead, enabling the PANet-style neck to perform effective and memory-efficient feature fusion across multiple scales.

% \begin{table}[h!]
% \centering
% \small
% \caption{Comparison of Styolod model variants under FP32, INT8, and INT4 quantization. Latency measured on target MCU.}
% \label{tab:styolod_variants}
% \rowcolors{2}{gray!10}{white}
% \begin{tabular}{l c c c c c}
% \hline
% \rowcolor{gray!20}
% \textbf{Model} & \textbf{Params (M)} & \textbf{Latency (ms)} & \multicolumn{3}{c}{\textbf{mAP (0.50:0.95)}} \\
% \rowcolor{gray!20}
%  &  &  & \textbf{FP32} & \textbf{INT8} & \textbf{INT4} \\
% \hline

\begin{table}[h!]
\centering
\small
\caption{Neck alignment between STYOLONano backbone and YOLOX-Nano neck.}
\begin{tabular}{lccc}
\hline
\textbf{Output} & \textbf{Stride} & \textbf{Backbone Ch.} & \textbf{Proj. Ch.} \\
\hline
\texttt{dark3} & 8  & 128 & 64  \\
\texttt{dark4} & 16 & 256 & 128 \\
\texttt{dark5} & 32 & 512 & 256 \\
\hline
\end{tabular}
\label{tab:neck_alignment}
\end{table}

\subsection{Learning Rate Optimization}
To enhance training stability and accelerate convergence, a layer-wise learning rate scaling strategy was adopted for different components of the STYOLO architecture. The backbone was trained with a reduced learning rate of $0.2\times$ the base value to preserve pretrained feature representations, while the neck was updated using $0.8\times$ the base learning rate to facilitate effective feature aggregation. The detection head, being randomly initialized and requiring faster adaptation, was trained with the full base learning rate ($1.0\times$).

This differential learning rate scheme, inspired by prior work on layer-wise optimization in object detectors~\cite{ge2021yolox, glenn2021yolov5}, enables a balance between stability in lower layers and rapid learning in higher layers. Empirically, this approach improved both convergence speed and final detection accuracy, increasing the mAP of \emph{STYOLO-Nano} from $21.32$ to $26.25$ on the MS-COCO dataset.

\subsection{RAM-Efficient Projection Layer} 
\label{subsec:projection_layer}
Memory profiling of \emph{STYOLOMicro} on the STM32N6 shows a RAM usage of 4.26~MB, higher than competing models such as YOLOv5n~\cite{Jocher_YOLOv5_by_Ultralytics_2020} and YOLOv8n~\cite{ultralytics2023yolov8}. Reducing RAM is crucial for MCU/NPU deployment, as N6 performance drops sharply beyond 4~MB due to external memory dependence. Layer-wise analysis identifies the stem layer’s final convolution (1$\times$1, 8→32) as the main contributor, driven by large feature maps and channel count.

To mitigate this, we modify the STResNet backbone by adding a lightweight \textit{projection layer}. The third convolution in the stem now outputs 32 instead of 64 channels, followed by a parallel 1$\times$1 projection (32→64) operating on smaller feature maps, as shown in Table~\ref{tab:proj_layer}. This balances efficiency and expressivity—reducing memory footprint while preserving channel diversity. Consequently, RAM usage drops from 4.26~MB to 2.46~MB with less than 0.5\% mAP degradation.

\begin{table}[h]
\centering
\caption{Comparison of original vs. modified Conv3 with projection layer.}
\label{tab:proj_layer}
\begin{tabular}{c|c}
\hline
{Original Conv3} & $64 \;\rightarrow\; 64$ \\ \hline
{Modified Conv3} & $64 \;\rightarrow\; 32$ \\ \hline
{Projection (1$\times$1)} & $32 \;\rightarrow\; 64$ \\ \hline
\end{tabular}
\end{table}

\section{Results} 
\label{lab:results}
\subsection{STResNet}

\begin{table*}[h!]
\centering
\small
\caption{Comparison of ResNet-18 with STResNet MCU variants, accuracy on ImageNet \cite{he2016deep}, performance data on STM32N6 board [*- too large to fit on internal RAM].}

\label{tab:resnet18_vs_stresnet}
\rowcolors{2}{gray!10}{white}
\begin{tabular}{l c c c c c c}
\hline
\rowcolor{gray!10}
\textbf{Model} & \textbf{Params (M)} & \textbf{Top-1 Acc. (\%)} & \textbf{Acc. Drop (\%)} & \textbf{Size Red.} & \textbf{Latency (ms)} & \textbf{RAM (MB)} \\
\hline

ResNet-18* \cite{he2016deep} & 11.68 & 70.5 & 0.0 & $1.0\times$ & - & - \\

STResNetTiny & 3.99 & 71.6 & $+1.1$ & $3\times$ & 21.29  & 1.39  \\

STResNetMilli & 3.00 & 70.0 & $-0.5$ & $3.89\times$ & 18.29  & 1.39  \\

STResNetMicro & 1.50 & 66.7 & $-3.8$ & $7.8\times$ & 14.36 & 0.882 \\

STResNetNano & 0.95 & 58.8 & $-11.7$ & $12.3\times$ & 10.91 & 0.833 \\

% \rowcolor{green!10}
STResNetPico & 0.62 & 48.8 & $-21.7$ & $18.8\times$ & 8.24 & 0.833 \\

\hline
\end{tabular}
\end{table*}

Table~\ref{tab:resnet18_vs_stresnet} shows performance of STResNet models on imagenet dataset along with Flash, RAM and latency requirments measured on STM32N6~\cite{stmicroelectronics2024n6} Neural Art NPU. The STResNet family demonstrates a strong trade-off between accuracy and efficiency on MCU hardware. Compared to ResNet-18, STResNet variants achieve up to 18.8× model size reduction and 2.6× faster latency on the STM32N6 board. While accuracy gradually decreases with smaller models, STResNetTiny even surpasses ResNet-18 by +1.1\%, showing that the architecture effectively balances compactness and performance for edge deployment.

All the models are trained on ImageNet~\cite{deng2009imagenet} dataset using default timm~\cite{rw2019timm} training pipeline for 300 epochs. Each model is benchmarked on STM32N6 using ST Edge AI Developer Cloud~\cite{stmicroelectronics2024n6} to get Flash, RAM and latency values.

Table \ref{tab:light_models_sota} presents a comparative analysis of several lightweight classification models with fewer than 4 M parameters on the ImageNet-1K dataset \cite{deng2009imagenet}. The proposed \emph{STResNet} family demonstrates competitive or superior accuracy compared to state-of-the-art handcrafted models such as MobileNet \cite{howard2017mobilenets, sandler2018mobilenetv2, howard2019mobilenetv3}, ShuffleNet \cite{zhang2018shufflenet, ma2018shufflenetv2}, and SqueezeNet \cite{iandola2016squeezenet}. Specifically, \emph{STResNetTiny} achieves 71.6\% Top-1 accuracy with 3.99 M parameters, closely matching MobileNetV2-1.0 while exhibiting lower RAM usage ($1.39$ vs $2.01$ MB and latency (21.3 ms vs. 22.4 ms). The mid-sized \emph{STResNetMilli} attains 70.0\% accuracy with 3.0 M parameters—outperforming MobileNetV1-0.75 and ShuffleNetV2-1.0$\times$ by 1.6–2.6\% absolute Top-1 accuracy at comparable or lower latency. The compact \emph{STResNetMicro} reaches 66.7\% accuracy with only 1.5 M parameters, exceeding ShuffleNetV2-0.5$\times$ by 5.7\% and demonstrating a favorable trade-off between accuracy and model size. Even the smallest variant, \emph{STResNetNano}, maintains 58.8\% accuracy at under 1 M parameters—comparable to SqueezeNet 1.1 but with significantly lower latency ($10.9$ ms vs. $119.9$ ms). \emph{STResNetMicro} has the best accuracy/size/latency tradeoff compared of all other comparable state of the art models.

\begin{table*}[h!]
\centering
\small % or \scriptsize / \footnotesize
\caption{Comparison of lightweight models on ImageNet-1K \cite{deng2009imagenet} with $<4$M parameters. Accuracy is reported on FP32 models; RAM, latency, and INT8 performance are measured on STM32N6~\cite{stmicroelectronics2024n6}}
\label{tab:light_models_sota}
\rowcolors{2}{gray!10}{white}
\begin{tabular}{l c c c c}
\hline
\rowcolor{gray!10}
\textbf{Model} & \textbf{Params (M)} & \textbf{Top-1 Acc. (\%)} & \textbf{RAM (MB)} & \textbf{Latency (ms)} \\
\hline

MobileNetV1-1.00 \cite{howard2017mobilenets} & 4.20 & 70.6 & 1.53 & 20.76 \\
\rowcolor{orange!10}
MobileNetV3-large-0.75 \cite{howard2019mobilenetv3} & 4.00 & 73.3 & 1.38 & 36.46 \\
\rowcolor{green!10}
\textbf{STResNetTiny} & 3.99 & 71.6 & 1.39 & 21.29 \\
MobileNetV2-1.00 \cite{howard2017mobilenets} & 3.50 & 71.8 & 2.01 & 22.44 \\
\hline

\rowcolor{green!10}
\textbf{STResNetMilli} & 3.00 & 70.0 & 1.39 & 18.29 \\

MobileNetV1-0.75 \cite{howard2017mobilenets} & 2.60 & 68.4 & 1.29 & 11.50 \\
MobileNetV3-small-1.0 \cite{howard2019mobilenetv3} & 2.53 & 67.4 & 1.579 & 54.35 \\
ShuffleNetV2 1.0$\times$ \cite{ma2018shufflenetv2} & 2.30 & 69.4 & 0.738 & 34.15 \\
MobileNetV2-0.5 \cite{sandler2018mobilenetv2} & 2.00 & 65.4 & 1.24 & 11.51 \\
MobileNetV3-small-0.75 \cite{howard2019mobilenetv3} & 1.99 & 65.4 & 1.579 & 33.12 \\
\rowcolor{orange!10}
ShuffleNetV1 1.0$\times$ \cite{zhang2018shufflenet} & 1.87 & 68.13 & 0.628 & 15.54 \\
MobileNetV2-0.35 \cite{sandler2018mobilenetv2} & 1.70 & 60.3 & 0.90 & 10.39 \\
\rowcolor{green!10}
\textbf{STResNetMicro} & 1.50 & 66.7 & 0.882 & 14.36 \\
\hline

ShuffleNetV2 0.5$\times$ \cite{ma2018shufflenetv2} & 1.40 & 61.0 & 0.735 & 8.54 \\
MobileNetV1-0.5 \cite{howard2017mobilenets} & 1.30 & 63.7 & 0.574 & 8.09 \\
SqueezeNet 1.0 \cite{iandola2016squeezenet} & 1.25 & 57.5 & 8.75 & 119.97 \\
SqueezeNet 1.1 \cite{iandola2016squeezenet} & 1.24 & 58.2 & 0.785 & 9.46 \\
\rowcolor{green!10}
\textbf{STResNetNano} & 0.95 & 58.8 & 0.833 & 10.91 \\
\hline

\rowcolor{orange!10}
MobileNetV1-0.25 \cite{howard2017mobilenets} & 0.50 & 50.6 & 0.383 & 3.88 \\
\rowcolor{green!10}
\textbf{STResNetPico} & 0.60 & 48.8 & 0.833 & 8.24 \\
\hline
\end{tabular}
\end{table*}

Another key aspect is the quantization-friendliness of compact models such as MobileNet~\cite{howard2017mobilenets, sandler2018mobilenetv2, howard2019mobilenetv3}, SqueezeNet~\cite{iandola2016squeezenet}, EfficientNet~\cite{tan2019efficientnet}, and ShuffleNet~\cite{ma2018shufflenetv2, zhang2018shufflenet}. Many of these networks suffer notable accuracy degradation under ultra-low-bit quantization (<8-bit), as shown in prior studies~\cite{li2021brecq, park2020profit}. SqueezeNet and EfficientNet are often excluded from such benchmarks due to their reliance on specialized layers that complicate quantization, typically requiring quantization-aware training (QAT) to recover accuracy.
In contrast, STResNet, derived from the regular ResNet~\cite{he2016deep} architecture, is inherently quantization-friendly. Its uniform convolutional structure and absence of exotic operators make it well-suited for MCU and NPU deployment, where low-bit quantization is essential. This design ensures strong performance in scenarios demanding both compactness and quantization efficiency.

\subsection{STYOLO}
Object detection models are trained on MS COCO~\cite{coco_test} dataset
for 300 epochs using YOLOX~\cite{ge2021yolox} training pipeline with the customization as explained in~ section \ref{sec:styolo}. 

\label{sec:styolo_results}

Table~\ref{tab:small_obj_det_n6} summarizes the performance comparison of various lightweight object detection models on the STM32N6 (Benchmarked at 320~px input resolution). The results demonstrate that the proposed \emph{STYOLO} family consistently achieves a favorable balance between accuracy, latency, and memory efficiency compared to existing compact detectors such as YOLOv5n~\cite{Jocher_YOLOv5_by_Ultralytics_2020}, YOLOv8n~\cite{ultralytics2023yolov8}, and YOLOX-nano~\cite{ge2021yolox}. \emph{STYOLOMicro} achieves a mAP of 30.54\% with only 1.69~M parameters outperforming YOLOv5n (+2.54~mAP) while maintaining comparable RAM usage (2.46~MB vs. 2.11~MB). Although YOLOv8n reaches a higher accuracy (35.60~mAP), it does so using custom Ultralytics training pipeline which features a lot of mAP improvement techniques. \emph{STYOLOMilli} is 2 mAP lower compared to YOLO8n using ReLU activation. We believe that using our backbone with YOLOv8n training pipeline, we can improve the accuracy of Ultralytics models as well and it is future work of this paper. Similarly, \emph{STYOLOMilli} and \emph{STYOLOTiny} models surpass their YOLOX-Tiny counterparts in accuracy (by +0.85~mAP and +2.65~mAP respectively) with lower parameter counts and improved scalability across hardware tiers.

\begin{table*}[h!]
\centering
\small
\caption{Comparison of Small Object Detection Models and Benchmarking on STM32N6 at 320px, [*- too large to fit on internal RAM]}
\label{tab:small_obj_det_n6}

\begin{tabular}{l c c c c c c c}

\hline

\multirow{2}{*}{\textbf{Model}} & 
\multirow{2}{*}{\textbf{Resolution}} & 
\multirow{2}{*}{\textbf{Activation}} & 
\multirow{2}{*}{\textbf{mAP}} & 
\multirow{2}{*}{\textbf{Params (M)}} & 
\multicolumn{3}{c}{\textbf{STM32N6 Benchmark} (320px)} \\
\cline{6-8}
& & & & & \textbf{Latency (ms)} & \textbf{Flash (MB)} & \textbf{RAM (MB)} \\
\hline
\rowcolor{orange!10}
YOLO8n       & 640 & SiLU & 37.30 & 3.20 & 40.98  & 3.28  & 2.17 \\
\rowcolor{orange!10}
YOLO8n       & 640 & ReLU & 35.60 & 3.20 & 40.98  & 3.28  & 2.17 \\
YOLOX-Tiny*  & 416 & ReLU & 32.80 & 5.06 & -  & -  & - \\

\rowcolor{green!10}
STYOLOMilli  & 640 & ReLU & 33.65 & 3.25 & 76.90 & 3.49 & 3.49 \\
\hline
\rowcolor{orange!10}
YOLO5n       & 640 & SiLU & 28.00 & 1.90 & 35.98  & 2.83  & 2.11 \\
\rowcolor{green!10}
STYOLOMicro  & 640 & ReLU & 30.54 & 1.69 & 47.32  & 2.00  & 2.46 \\
\hline
\rowcolor{green!10}
STYOLONano   & 640 & ReLU & 26.25 & 1.13 & 41.37  & 1.47  & 2.46 \\

\rowcolor{green!10}
STYOLOPico   & 640 & ReLU & 20.54 & 0.74 & 28.53  & 1.07  & 0.75 \\

\hline
\rowcolor{green!10}
STYOLOPico   & 416 & ReLU & 18.14 & 0.74 & 28.53  & 1.07  & 0.75 \\

\rowcolor{orange!10}
YOLOX-Nano   & 416 & ReLU & 23.80 & 1.08 & 139.30 & 1.31  & 2.41 \\
\rowcolor{green!10}
STYOLONano   & 416 & ReLU & 23.60 & 1.13 & 41.37  & 1.47  & 2.46 \\
\hline
\end{tabular}
\end{table*}

\begin{table*}[h!]
\centering
\small
\caption{STYOLO-RAM improvement using projection layer}
\label{tab:ram_improvement_projection_layer}

\begin{tabular}{l c c c c c c c}
\hline
\multirow{2}{*}{\textbf{Model}} & 
\multirow{2}{*}{\textbf{Resolution}} & 
\multirow{2}{*}{\textbf{Activation}} & 
\multirow{2}{*}{\textbf{mAP}} & 
\multirow{2}{*}{\textbf{Params (M)}} & 
\multicolumn{3}{c}{\textbf{STM32N6 Benchmark} (320px)} \\
\cline{6-8}
& & & & & \textbf{Latency (ms)} & \textbf{Flash (MB)} & \textbf{RAM (MB)} \\
\hline

\rowcolor{orange!10}
STYOLOMicro & 640 & ReLU & 30.54 & 1.69 & 47.32 & 2.00 & 4.26 \\

\rowcolor{green!10}
{STYOLOMicro (Proj)} & 
{640} & 
{ReLU} & 
\textbf{30.12} & 1.69 & {42.99} & 2.00 & \textbf{2.46} \\

\hline
\end{tabular}
\end{table*}

At smaller configurations, \emph{STYOLONano} achieves competitive accuracy (23.6~mAP), slightly below YOLOX-nano (23.8~mAP). Interestingly, YOLOX-nano is $3\times$ slower than STYOLOTiny, highlighting that certain specialized operators in compact models are not MCU/NPU-friendly. The \emph{STYOLOPico} variant defines the lower bound of the design space with only 0.74~M parameters, offering a deployable option for sub-1~MB memory targets. Using ReLU activations instead of SiLU further simplifies fixed-point deployment while maintaining accuracy, reinforcing the design’s suitability for embedded applications. As shown in Table~\ref{tab:ram_improvement_projection_layer}, the proposed RAM-efficient projection layer reduces runtime memory in STYOLOMicro from 4.26~MB to 2.46~MB (a 42\% decrease) with comparable accuracy (30.54~mAP~$\rightarrow$~30.12~mAP) and slightly better latency (47.32~ms~$\rightarrow$~42.99~ms), validating its effectiveness in minimizing feature map size without loss of representation quality.

\subsection{Ultralytics Experiments} 
\label{subsec:ultralytics_experiments}
To verify that the proposed methodology—integrating optimized pre-trained backbones with existing neck and head architectures—generalizes across different training environments, we extended our experiments to the Ultralytics YOLOv11 framework. Specifically, the STResNetMilli and STResNetMicro models, pre-trained on the ImageNet-1K dataset, were attached to channel-adjusted neck and head modules from YOLOv11 and subsequently fine-tuned on the full MS-COCO dataset. The resulting STResNetMicro-YOLOv11 model achieved a performance level comparable to the original YOLOv11n. When trained under the same training environment as Ultralytics  and learning rate optimization settings described earlier, the proposed model closely matched the performance of YOLOv11n as shown in Table~\ref{tab:ultralytics_models}, validating the adaptability and robustness of our approach across diverse training pipelines.It is also evident from the table that STResNet backbone performed almost similar to YOLO11n at lower resolution training which is very cruicial for MCU and NPU devices given the memory constraints.

\begin{table}[h!]
\centering
\small
\caption{STResNet backbone performance with Ultralytics pipeline.}
\label{tab:ultralytics_models}
\begin{tabular}{lccc}
\hline
\rowcolor{gray!10}
\textbf{Backbone} & \textbf{mAP(50:95)} & \textbf{Params} \\
\rowcolor{gray!10}
\textbf{/Head} & \textbf{/Resolution} & \textbf{(M)} \\
\hline
YOLO11n/YOLOv11n & 22.60/224  & 2.60  \\
STResNetMicro/YOLOv11n & 22.60/224  & 2.90  \\
YOLO11n/YOLOv11n & 39.50/640  & 2.60  \\
STResNetMicro/YOLOv11n & 38.60/640 & 2.90  \\
STResNetMilli/YOLOv11n & 40.50/640  & 4.46  \\
STResNetMicro/YOLOv11s & 42.01/640 & 5.50  \\
YOLO11s & 47.00/640  & 9.4  \\

\hline
\end{tabular}
\label{tab:neck_alignment}
\end{table}

\section{Conclusion}
\label{sec:conclusion}

In this work, we presented a family of lightweight, hardware-efficient neural networks tailored for MCU and NPU deployment. The proposed \emph{STResNet} combines low-rank layer decomposition with NAS-guided channel optimization (CompressNAS), forming a simplified yet effective ResNet variant that achieves 3–12$\times$ compression while maintaining competitive accuracy. Unlike handcrafted models such as MobileNet or EfficientNet, STResNet avoids specialized operators, ensuring quantization stability and efficient fixed-point execution.

Building on this compact backbone, we introduced the \emph{STYOLO} series of object detectors that integrate STResNet within a YOLOX-style framework. Extensive experiments and hardware benchmarks on the STM32N6 Neural Art NPU show that STYOLO models deliver superior accuracy–efficiency trade-offs compared to YOLOv5n, YOLOv8n, and YOLOX-Nano. The proposed RAM-efficient projection layer further reduced memory usage by 42\% and improved latency by 9\%, demonstrating its effectiveness for edge deployment.

\section{Future Work}

In future, we plan to extend this framework in three directions: (1) integrate mixed-precision quantization into the CompressNAS search to jointly optimize bit-width and rank for better accuracy–efficiency balance; (2) adapt the STResNet backbone for multi-task edge vision tasks such as segmentation and keypoint detection; and (3) develop cross-hardware adaptive compression methods to automatically tune decomposition and channel scaling for diverse MCU, NPU, and DSP architectures.

% \section*{Acknowledgment} This research was supported by anonymous funding sources.

% ---- Bibliography ----
%%
\bibliographystyle{unsrt}
\bibliography{references}

% \newpage
\null
\thispagestyle{empty}
\newpage

\input{appendix}

\end{document}

%% file: appendix.tex
\begin{table}[h!]
\centering
\small
\caption{STResNet-Pico Architecture }
\label{tab:stresnet_pico_compact}
\begin{tabular}{l c c}
\hline
\textbf{Block / Layer} & \textbf{In $\to$ Out} & \textbf{Conv Layers (Inner Channels)} \\
\hline
\textbf{Stem} & 3 $\to$ 64 & 1x1/3→3, 7x7/3→8, 1x1/8→32 \\ & & \textbf{proj.1x1/32→64} \\
\hline
\textbf{Layer1} & 64 $\to$ 64 & Block1: 1x1/64→24, 3x3/24, 1x1/24→64; 1x1/64→16, 3x3/16, 1x1/16→64 \\
& & Block2: 1x1/64→24, 3x3/24, 1x1/24→64; 1x1/64→8, 3x3/8, 1x1/8→64 \\
\hline
\textbf{Layer2} & 64 $\to$ 128 & Block1: 1x1/64→24, 3x3/24 s2, 1x1/24→128; 1x1/128→8, 3x3/8, 1x1/8→128 \\
& & Block2: 1x1/128→8, 3x3/8, 1x1/8→128; 1x1/128→8, 3x3/8, 1x1/8→128 \\
\hline
\textbf{Layer3} & 128 $\to$ 256 & Block1: 1x1/128→8, 3x3/8 s2, 1x1/8→256; 1x1/256→8, 3x3/8, 1x1/8→256 \\
& & Block2: 1x1/256→8, 3x3/8, 1x1/8→256; 1x1/256→8, 3x3/8, 1x1/8→256 \\
\hline
\textbf{Layer4} & 256 $\to$ 512 & Block1: 1x1/256→8, 3x3/8 s2, 1x1/8→512; 1x1/512→8, 3x3/8, 1x1/8→512 \\
& & Block2: 1x1/512→8, 3x3/8, 1x1/8→512; 1x1/512→8, 3x3/8, 1x1/8→512 \\
\hline
\end{tabular}
\end{table}

\begin{table}[h!]
\centering
\small
\caption{STResNet-Tiny Architecture}
\label{tab:stresnet_tiny_compact}
\begin{tabular}{l c c}
\hline
\textbf{Block / Layer} & \textbf{In $\to$ Out} & \textbf{Conv Layers (Inner Channels)} \\
\hline
\textbf{Stem} & 3 $\to$ 64 & 1x1/3→3, 7x7 s2/3→16, 1x1/16→32 \\ & & \textbf{proj.1x1/32→64} \\ 
\hline
\textbf{Layer1} & 64 $\to$ 64 & Block1: 3x3/64→64, 3x3/64→64 \\
& & Block2: 3x3/64→64, 3x3/64→64 \\
\hline
\textbf{Layer2} & 64 $\to$ 128 & Block1: 3x3 s2/64→128; 1x1/128→96, 3x3/96, 1x1/96→128 \\
& & Block2: 3x3/128→128; 1x1/128→80, 3x3/80, 1x1/80→128 \\
\hline
\textbf{Layer3} & 128 $\to$ 256 & Block1: 3x3 s2/128→256; 1x1/256→192, 3x3/192, 1x1/192→256 \\
& & Block2: 3x3/256→256; 1x1/256→96, 3x3/96, 1x1/96→256 \\
\hline
\textbf{Layer4} & 256 $\to$ 512 & Block1: 1x1/256→208, 3x3 s2/208→208, 1x1/208→512; \\ & &  1x1/512→88, 3x3/88, 1x1/88→512 \\
& & Block2: 1x1/512→192, 3x3/192, 1x1/192→512; \\ & & 1x1/512→112, 3x3/112, 1x1/112→512 \\
\hline
\end{tabular}
\end{table}

\begin{table}[h!]
\centering
\small
\caption{STResNet-Micro Architecture }
\label{tab:stresnet_micro_compact}
\begin{tabular}{l c c}
\hline
\textbf{Block / Layer} & \textbf{In $\to$ Out} & \textbf{Conv Layers (Inner Channels)} \\
\hline
\textbf{Stem} & 3 $\to$ 64 & 1x1/3→3, 7x7/3→8, 1x1/8→32 \\ & & \textbf{proj.1x1/32→64} \\
\hline
\textbf{Layer1} & 64 $\to$ 64 & Block1: 1x1/64→64, 3x3/64, 1x1/64→64; 1x1/64→64, 3x3/64, 1x1/64→64 \\
& & Block2: 1x1/64→64, 3x3/64, 1x1/64→64; 1x1/64→64, 3x3/64, 1x1/64→64 \\
\hline
\textbf{Layer2} & 64 $\to$ 128 & Block1: 1x1/64→40, 3x3/40 s2, 1x1/40→128; 1x1/128→32, 3x3/32, 1x1/32→128 \\
& & Block2: 1x1/128→88, 3x3/88, 1x1/88→128; 1x1/128→32, 3x3/32, 1x1/32→128 \\
\hline
\textbf{Layer3} & 128 $\to$ 256 & Block1: 1x1/128→88, 3x3/88 s2, 1x1/88→256; 1x1/256→72, 3x3/72, 1x1/72→256 \\
& & Block2: 1x1/256→80, 3x3/80, 1x1/80→256; 1x1/256→32, 3x3/32, 1x1/32→256 \\
\hline
\textbf{Layer4} & 256 $\to$ 512 & Block1: 1x1/256→80, 3x3/80 s2, 1x1/80→512; 1x1/512→8, 3x3/8, 1x1/8→512 \\
& & Block2: 1x1/512→72, 3x3/72, 1x1/72→512; 1x1/512→64, 3x3/64, 1x1/64→512 \\
\hline
\end{tabular}
\end{table}

%% file: references.bib
@String(CVPR  = {IEEE Conf. Comput. Vis. Pattern Recog.})

@String(ECCV  = {Eur. Conf. Comput. Vis.})

@String(NeurIPS = {Adv. Neural Inform. Process. Syst.})

@String(ICML  = {Int. Conf. Mach. Learn.})

@String(ICLR  = {Int. Conf. Learn. Represent.})

@String(CVPR  = {CVPR})

@String(ECCV  = {ECCV})

@String(NeurIPS = {NeurIPS})

@String(ICML  = {ICML})

@String(ICLR  = {ICLR})

@misc{coco_test,
  title = {COCO Detection Challenge},
  howpublished = {\url{https://codalab.lisn.upsaclay.fr/competitions/7384}},
    author = {},
    year={}
}

@software{yolo11_ultralytics,
  author = {Glenn Jocher and Jing Qiu},
  title = {Ultralytics YOLO11},
  version = {11.0.0},
  year = {2024},
  url = {https://github.com/ultralytics/ultralytics},
  orcid = {0000-0001-5950-6979, 0000-0003-3783-7069},
  license = {AGPL-3.0}
}

@misc{lin2014microsoft,
  author = {Lin, Tsung-Yi and Maire, Michael and Belongie, Serge and Bourdev, Lubomir and Girshick, Ross and Hays, James and Perona, Pietro and Ramanan, Deva and Zitnick, C. Lawrence and Dollár, Piotr},
  description = {[1405.0312] Microsoft COCO: Common Objects in Context},
  note = {cite arxiv:1405.0312Comment: 1) updated annotation pipeline description and figures; 2) added new  section describing datasets splits; 3) updated author list},
  title = {Microsoft COCO: Common Objects in Context},
  url = {http://arxiv.org/abs/1405.0312},
  year = {2014}
}

@inproceedings{deng2009imagenet,
  author = {Deng, Jia and Socher, R. and Fei-Fei, Li and Dong, Wei and Li, Kai and Li, Li-Jia},
  booktitle = {2009 IEEE Conference on Computer Vision and Pattern Recognition(CVPR)},
  description = {ImageNet: A large-scale hierarchical image database},
  doi = {10.1109/CVPR.2009.5206848},
  month = {06},
  pages = {248-255},
  title = {ImageNet: A large-scale hierarchical image database},
  url = {https://ieeexplore.ieee.org/abstract/document/5206848/},
  volume = {00},
  year = {2009}
}

@software{Jocher_YOLOv5_by_Ultralytics_2020,
author = {Jocher, Glenn},
doi = {10.5281/zenodo.3908559},
license = {AGPL-3.0},
month = may,
title = {{YOLOv5 by Ultralytics}},
url = {https://github.com/ultralytics/yolov5},
version = {7.0},
year = {2020}
}

@article{relu,
  title={Deep Learning using Rectified Linear Units (ReLU)},
  author={Abien Fred Agarap},
  journal={arXiv preprint arXiv:1803.08375},
  year={2018}
}

@inproceedings{howard2017mobilenets,
  title={MobileNets: Efficient Convolutional Neural Networks for Mobile Vision Applications},
  author={Howard, Andrew G and Zhu, Menglong and Chen, Bo and Kalenichenko, Dmitry and Wang, Weijun and Weyand, Tobias and Andreetto, Marco and Adam, Hartwig},
  year={2017},
  booktitle={arXiv preprint arXiv:1704.04861}
}

@inproceedings{howard2019mobilenetv3,
  title={Searching for MobileNetV3},
  author={Howard, Andrew and Sandler, Mark and Chu, Grace and Chen, Liang-Chieh and Chen, Bo and Tan, Mingxing and Wang, Weijun and Zhu, Yukun and Pang, Ruoming and Vasudevan, Vijay and Le, Quoc V and Adam, Hartwig},
  booktitle={Proceedings of the IEEE/CVF International Conference on Computer Vision},
  pages={1314--1324},
  year={2019}
}

@inproceedings{sandler2018mobilenetv2,
  title={MobileNetV2: Inverted Residuals and Linear Bottlenecks},
  author={Sandler, Mark and Howard, Andrew and Zhu, Menglong and Zhmoginov, Andrey and Chen, Liang-Chieh},
  booktitle={Proceedings of the IEEE Conference on Computer Vision and Pattern Recognition},
  pages={4510--4520},
  year={2018}
}

@inproceedings{ma2018shufflenetv2,
  title={ShuffleNet V2: Practical Guidelines for Efficient CNN Architecture Design},
  author={Ma, Ningning and Zhang, Xiangyu and Zheng, Hai-Tao and Sun, Jian},
  booktitle={Proceedings of the European Conference on Computer Vision (ECCV)},
  pages={116--131},
  year={2018}
}

@inproceedings{zhang2018shufflenet,
  title={ShuffleNet: An Extremely Efficient Convolutional Neural Network for Mobile Devices},
  author={Zhang, Xiangyu and Zhou, Xinyu and Lin, Mengxiao and Sun, Jian},
  booktitle={Proceedings of the IEEE Conference on Computer Vision and Pattern Recognition},
  pages={6848--6856},
  year={2018}
}

@inproceedings{iandola2016squeezenet,
  title={SqueezeNet: AlexNet-level accuracy with 50x fewer parameters and <0.5MB model size},
  author={Iandola, Forrest N and Han, Song and Moskewicz, Matthew W and Ashraf, Khalid and Dally, William J and Keutzer, Kurt},
  booktitle={arXiv preprint arXiv:1602.07360},
  year={2016}
}

@inproceedings{li2021brecq,
  title     = {BRECQ: Pushing the Limit of Post-Training Quantization by Block Reconstruction},
  author    = {Li, Yunfei and Dong, Zhen and Yang, Hanting and Liu, Shaokai and Hu, Zihan and Wang, Yunhe},
  booktitle = {International Conference on Learning Representations (ICLR)},
  year      = {2021}
}

@inproceedings{park2020profit,
  title     = {ProFit: Progressive Filter Pruning Globally at Fine-grained Level},
  author    = {Park, Jongsoo and Li, Sheng and Kim, Wonje and Lin, Shuaiwen and Keckler, Stephen W.},
  booktitle = {International Conference on Machine Learning (ICML)},
  year      = {2020},
  pages     = {7590--7600}
}

@inproceedings{ge2021yolox,
  title={YOLOX: Exceeding YOLO Series in 2021},
  author={Ge, Zheng and Liu, Songtao and Wang, Feng and Li, Zeming and Sun, Jian},
  booktitle={arXiv preprint arXiv:2107.08430},
  year={2021}
}

@misc{glenn2021yolov5,
  title={YOLOv5 by Ultralytics},
  author={Jocher, Glenn and others},
  howpublished={\url{https://github.com/ultralytics/yolov5}},
  year={2021}
}

@inproceedings{he2016deep,
  title={Deep residual learning for image recognition},
  author={He, Kaiming and Zhang, Xiangyu and Ren, Shaoqing and Sun, Jian},
  booktitle={CVPR},
  year={2016}
}

@inproceedings{tan2019efficientnet,
  title={EfficientNet: Rethinking model scaling for convolutional neural networks},
  author={Tan, Mingxing and Le, Quoc V.},
  booktitle={ICML},
  year={2019}
}

@article{han2015deep,
  title={Deep compression: Compressing deep neural networks with pruning, trained quantization and Huffman coding},
  author={Han, Song and Mao, Huizi and Dally, William J.},
  journal={ICLR},
  year={2016}
}

@inproceedings{jacob2018quantization,
  title={Quantization and training of neural networks for efficient integer-arithmetic-only inference},
  author={Jacob, Benoit and others},
  booktitle={CVPR},
  year={2018}
}

@inproceedings{denton2014exploiting,
  title={Exploiting linear structure within convolutional networks for efficient evaluation},
  author={Denton, Emily and others},
  booktitle={NeurIPS},
  year={2014}
}

@inproceedings{kim2015compression,
  title={Compression of deep convolutional neural networks for fast and low power mobile applications},
  author={Kim, Yunchao and Park, Eunwoo and Yoo, Sungjoo},
  booktitle={ICLR},
  year={2016}
}

@inproceedings{lebedev2014speeding,
  title={Speeding-up convolutional neural networks using fine-tuned CP-decomposition},
  author={Lebedev, Vadim and Lempitsky, Victor},
  booktitle={ICLR},
  year={2015}
}

@inproceedings{elsken2019neural,
  title={Neural architecture search: A survey},
  author={Elsken, Thomas and Metzen, Jan Hendrik and Hutter, Frank},
  booktitle={Journal of Machine Learning Research},
  year={2019}
}

@inproceedings{wu2019fbnet,
  title={FBNet: Hardware-aware efficient convnet design via differentiable neural architecture search},
  author={Wu, Bichen and others},
  booktitle={CVPR},
  year={2019}
}

@misc{rw2019timm,
  author       = {Ross Wightman},
  title        = {PyTorch Image Models (timm)},
  year         = {2019},
  publisher    = {GitHub},
  howpublished = {\url{https://github.com/rwightman/pytorch-image-models}},
  doi          = {10.5281/zenodo.4414861}
}

@software{jocher2020yolov5,
  title        = {YOLOv5 by Ultralytics},
  author       = {Glenn Jocher and Ayush Chaurasia and Jing Qiu},
  year         = {2020},
  howpublished = {\url{https://github.com/ultralytics/yolov5}},
  note         = {Accessed: 2025-10-03}
}

@software{ultralytics2023yolov8,
  title        = {Ultralytics YOLOv8},
  author       = {Glenn Jocher and Ayush Chaurasia and Jing Qiu},
  year         = {2023},
  howpublished = {\url{https://github.com/ultralytics/ultralytics}},
  note         = {Version 8.0, Accessed: 2025-10-03}
}

@misc{stmicroelectronics2024n6,
  title={STM32N6 AI NPU platform for next-generation embedded vision},
  author={{STMicroelectronics}},
  year={2024},
  note={\url{https://www.st.com/en/microcontrollers-microprocessors/stm32n6.html}}
}

@inproceedings{zoph2018nasnet,
  title={Learning Transferable Architectures for Scalable Image Recognition},
  author={Zoph, Barret and Vasudevan, Vijay and Shlens, Jonathon and Le, Quoc V},
  booktitle={CVPR},
  year={2018}
}

@inproceedings{zoph2018learning,
  title={Learning transferable architectures for scalable image recognition},
  author={Zoph, Barret and Vasudevan, Vijay and Shlens, Jonathon and Le, Quoc V},
  booktitle={Proceedings of the IEEE Conference on Computer Vision and Pattern Recognition (CVPR)},
  pages={8697--8710},
  year={2018}
}

@inproceedings{tan2019mnasnet,
  title={Mnasnet: Platform-aware neural architecture search for mobile},
  author={Tan, Mingxing and Chen, Bo and Pang, Ruoming and Vasudevan, Vijay and Sandler, Mark and Howard, Andrew and Le, Quoc V},
  booktitle={Proceedings of the IEEE/CVF Conference on Computer Vision and Pattern Recognition (CVPR)},
  pages={2820--2828},
  year={2019}
}

@inproceedings{cai2019proxylessnas,
  title={ProxylessNAS: Direct neural architecture search on target task and hardware},
  author={Cai, Han and Zhu, Ligeng and Han, Song},
  booktitle={International Conference on Learning Representations (ICLR)},
  year={2019}
}
